\documentclass[journal]{IEEEtran}

\usepackage{booktabs}
\usepackage{amsmath}
\usepackage{graphicx}
\usepackage{url}

\begin{document}

\title{Vulnerable road user detection:\\ state-of-the-art and open challenges}

\author{Patrick Mannion
\thanks{P. Mannion was with the Department
of Computer Science and Applied Physics, Galway-Mayo Institute of Technology, Galway,
H91 T8NW Ireland e-mail: patrick.mannion@gmit.ie}
}

\maketitle

\begin{abstract}
Correctly identifying vulnerable road users (VRUs), e.g. cyclists and pedestrians, remains one of the most challenging environment perception tasks for autonomous vehicles (AVs). This work surveys the current state-of-the-art in VRU detection, covering topics such as benchmarks and datasets, object detection techniques and relevant machine learning algorithms. The article concludes with a discussion of remaining open challenges and promising future research directions for this domain. 
\end{abstract}


\IEEEpeerreviewmaketitle

\section{Introduction}

\IEEEPARstart{R}{esearch} and development work on autonomous vehicles (AVs) has accelerated in recent years, driven by technological advances in both hardware and software. However, a number of significant challenges must be addressed before widespread adoption of AVs will be possible. These include legal and regulatory issues, public perceptions of AV technologies, cost, accuracy and reliability of on-board sensing equipment, computational constraints and the limitations of current algorithms and systems for tasks such as environment perception, path planning and control.

\begin{figure}[t]
\centering
\includegraphics[width=\linewidth]{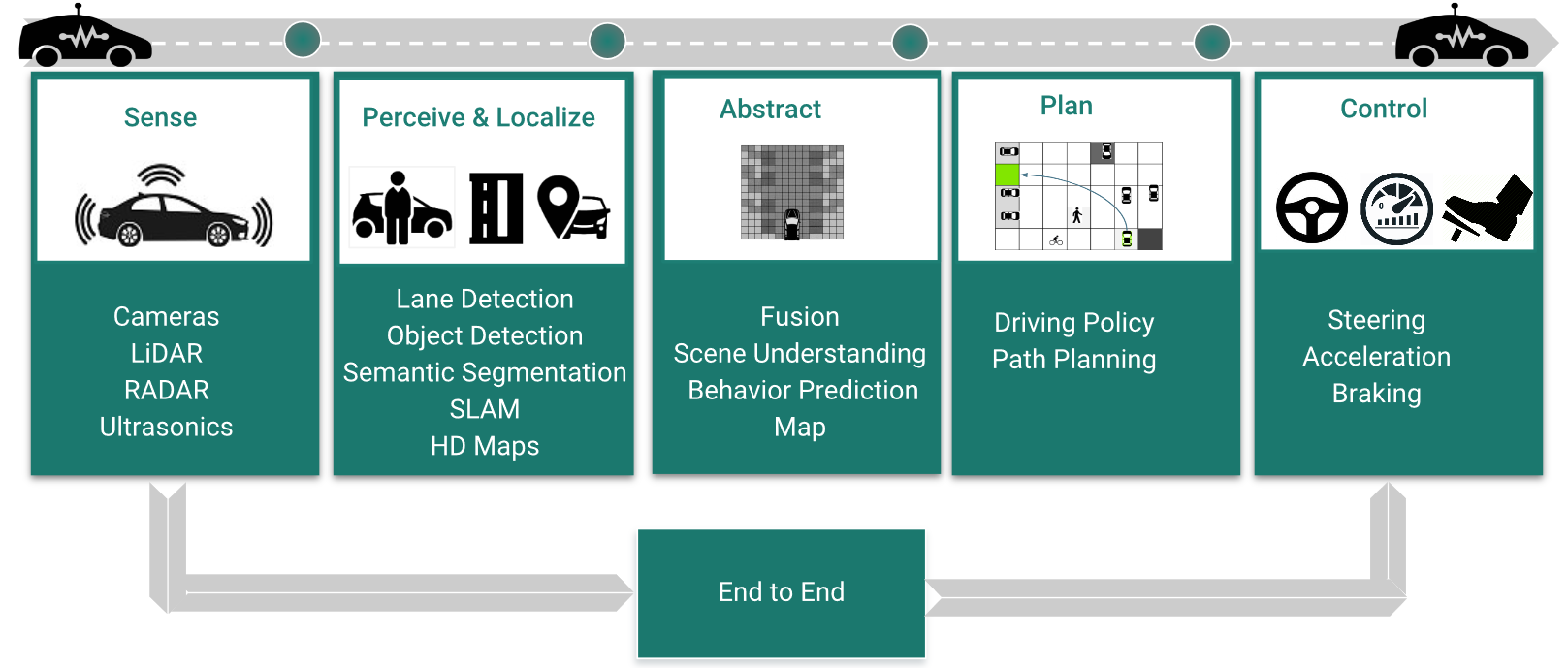}
\caption{Overview of autonomous driving tasks. Figure reproduced from \cite{Sobh2019Exploring}.}
\label{fig:ADflow}
\end{figure}

An overview of the main tasks which must be performed by an autonomous driving system is presented in Fig. \ref{fig:ADflow}. Environment perception is arguably one of the most important fields for future AV development; accurate perception of the environment allows automated driver assistance systems (ADASs) to make informed decisions during functions such as adaptive cruise control, lane changing, parking and obstacle and collision avoidance. Components of the environment which must be detected by an ego vehicle include: traffic signals and signs, road markings, lane and junction topology and other road users including vehicles, cyclists and pedestrians. 

Vulnerable road users (VRUs), such as pedestrians and especially cyclists, remain among the most challenging objects for AV perception systems to detect accurately \cite{fairley2017self}. These road users have little or protection during a collision when compared to vehicle occupants; this is reflected in the high proportion of VRUs among traffic fatality statistics. In recent years, traffic accidents have been the leading cause of death worldwide for people aged 18-29 \cite{who2015}. Approximately 1.25 million people died on roads worldwide in 2013; almost half of these fatalities were VRUs \cite{who2015}. A similar proportion of traffic fatalities during 2017 in the European Union were VRUs \cite{sajn2018general}. Therefore, improving VRU detection capabilities to human-level performance or above will be necessary over the coming years, if AVs are minimise the risk that they pose to VRUs and be accepted by legislators and the public.

This article aims to provide an overview of the current state-of-the-art in VRU detection, along with a discussion of the main open challenges and promising directions for future research. Other related surveys on environment perception for AVs \cite{bernini2014real,zhu2017overview,janai2017computer}, pedestrian detection \cite{benenson2014ten,zhang2016how} and perception benchmarks \cite{dollar2012pedestrian,braun2018EuroCity} may also be of interest to the reader. 

The next section of this paper provides a brief overview of the sensors which are used in AV perception, followed by Section \ref{sec:datasets} which discusses the main datasets which may be used to develop and test detection algorithms. Section \ref{sec:objdet} introduces related work on generating proposals for object detection, while Section \ref{sec:algorithms} presents an overview of the current best-performing algorithms for VRU detection. Section \ref{sec:openchallenges} discusses the main open challenges and some promising future research directions for VRU detection. Finally, Section \ref{sec:conclusion} concludes this paper with some closing remarks.

\section{Sensors for environment perception}
Sensor types which are currently used for AV environment perception include passive sensors such as visible spectrum (VS) and infrared (IR) cameras and active sensors such as Radio Detection and Ranging (RADAR), Light Detection and Ranging (LIDAR) and ultrasonic.

Cameras provide a wide field of view, although they have limited depth perception capabilities when used individually. VS cameras are among the cheapest and most widely used sensors, however their performance degrades significantly in low-light conditions as they work by capturing the light intensity values at each pixel \cite{janai2017computer}. IR cameras capture heat signatures using relative temperatures, so they are suitable for low-light conditions. Cameras may be utilised in stereo arrangements with a lateral offset to improve depth perception capabilities \cite{bernini2014real}.

Active sensors that emit signals and observe their reflections are better at providing range information than cameras, although they are more expensive. Accurate range information is important when determining the position in 3D space of a detected object in relation to the ego vehicle (i.e. when performing localisation). LIDAR may be used to capture highly accurate spatial data in the form of a 3D point cloud, but it is currently too expensive for widespread adoption on consumer vehicles.

Relying on one type of sensor alone may be problematic in certain driving conditions, therefore multiple types of sensors are typically used on AVs, combining the strengths and mitigating the weaknesses of each individual sensor type. The process of integrating information from multiple sensors is known as sensor fusion (see e.g. \cite{premebida2007lidar,ogawa2011pedestrian,han2015real,gonzalez2017onboard,zhang2018fusion}).

\section{Datasets and benchmarks for VRU detection}
\label{sec:datasets}

This section gives an overview of recent datasets and benchmarks which are most relevant for VRU detection. The public availability of datasets is one of the main factors which enabled the improvements in AV perception seen in recent years. As well as removing the requirement for researchers to have access to physical AV hardware, many of these datasets have dedicated benchmarking websites where new algorithms may be independently evaluated using a standard methodology. This allows prior results to be easily replicated and compared using online leaderboards. Most of these datasets focus on detection using VS cameras; some also provide synchronised LIDAR and camera frames. The highly accurate LIDAR data serves as a ground truth source, allowing the comparison of image-based detection algorithms against 3D point cloud data. 

The key features of the most important publicly available datasets are summarised in Table \ref{tab:datasets}. The datasets listed are EuroCity Persons (ECP) \cite{braun2018EuroCity}, CityPersons (CP) \cite{zhang2017citypersons}, Caltech \cite{dollar2012pedestrian}, KITTI \cite{geiger2012we} and the Tsinghua-Daimler Cyclist benchmark (TDC) \cite{li2016tdc}. EuroCity Persons is currently the most comprehensive publicly available dataset, as it has data from all seasons in both dry and weather conditions, from 31 cities across 12 countries. As a result ECP has much greater diversity than the other datasets surveyed, including samples of a wide variety of clothing types, weather types and lighting conditions. ECP is therefore currently the best overall choice for researchers wishing to develop and test new algorithms for VRU detection; this could be supplemented by other datasets such as Caltech to obtain a larger training set.

\begin{table}[h]
\caption{Comparison of publicly available VRU detection datasets}

\centering
$\begin{array}{ *{6}{c} }
\toprule
\text{Dataset} & \text{ECP \cite{braun2018EuroCity}} & \text{CP \cite{zhang2017citypersons}} & \text{Caltech \cite{dollar2012pedestrian}} & \text{KITTI \cite{geiger2012we}} & \text{TDC \cite{li2016tdc}} \\
\midrule
\text{\# seasons} & 4 & 3 & 1 & 1 & 1   \\ 
\text{weather}    & \text{dry \& wet} & \text{dry} & \text{dry} & \text{dry} & \text{dry}  \\
\text{lighting}   & \text{day \& night} & \text{day} & \text{day} & \text{day} & \text{day} \\
\text{\# images}   & 47337 & 5000 & 249884 & 14999 & 14674  \\
[1ex]
\bottomrule
\end{array}$
\label{tab:datasets}
\end{table}

\section{Object proposal generation}
\label{sec:objdet}
Objects detection from images is an important task in fields such as robotics and surveillance, as well as in environment perception for AVs. The object detection task consists of localisation (determining the position of an object in the image) and classification (determining which category an object belongs to) \cite{zhao2018object}. Traditional objection detection techniques try to identify regions of interest (ROIs) \cite{flohr2018vulnerable}, i.e. areas of an image which may contain the types of objects which must be detected. ROIs may be generated using a sliding window approach, by shifting a detector over an image at different scales, although exhaustive searches of this manner can be computationally expensive. Algorithms such as selective search \cite{uijlings2013selective} may be used to generate ROI proposals with less computational overhead.

\begin{figure}[h]
\centering
\includegraphics[width=0.5\linewidth]{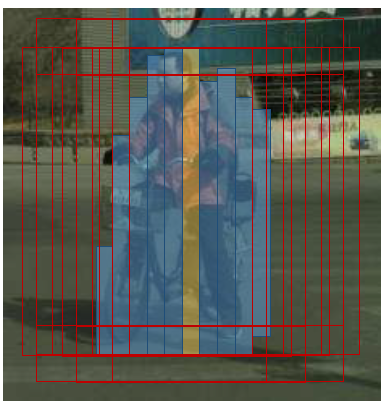}
\caption{Sample stixel-based bounding box proposals for a cyclist. Figure reproduced from \cite{flohr2018vulnerable}.}
\label{fig:stixel}
\end{figure}

As an alternative to rectangular ROIs, stixels \cite{badino2009stixel} may also be used for generating object proposals from images. A sample stixel-based proposal is shown in Fig. \ref{fig:stixel}. Stixels are a medium-level representation, intended to bridge the gap between individual pixels and actual objects. Individual stixels have a specific fixed width and variable height in pixels, are defined by their 3D position relative to the camera and stand vertically on the ground plane. Stixels are therefore suitable for generating proposals for objects which are mostly vertical (e.g. VRUs and vehicles).

\begin{figure}[h]
\centering
\includegraphics[width=0.7\linewidth]{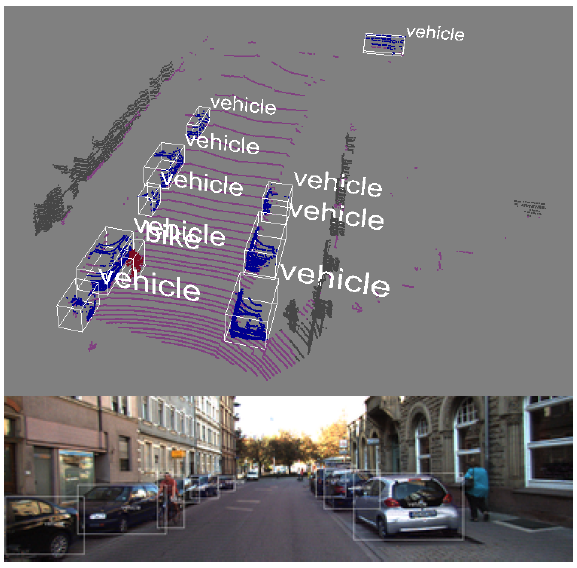}
\caption{Sample LIDAR-based bounding box proposals using point clustering. Figure reproduced from \cite{braun2016pose}.}
\label{fig:lidarprop}
\end{figure}

Object proposals may also be generated from LIDAR point cloud data, as illustrated in Fig. \ref{fig:lidarprop}. One commonly used approach is to first remove all the points corresponding to the ground surface and then to group the remaining points using a nearest-neighbour clustering algorithm (e.g. a kd-tree search structure) \cite{braun2016pose}. Finally a 3D bounding box representing each LIDAR point cluster is projected onto the image plane to generate 2D object proposals.

\section{VRU detection algorithms}
\label{sec:algorithms}
Once good quality object proposals have been generated, a classification algorithm must be run on each of the proposals. Many of the most prominent classification methods use machine learning (ML), a process which allows a computer program to learn to improve its performance at a task with increased experience \cite{Mitchell1997Machine}. ML algorithms for AV perception are generally trained on a dataset with labelled examples like those in Section \ref{sec:datasets}. The goal for a designer is to develop an algorithm which can successfully generalise what it has learned on the training dataset to classify new, unseen examples (either proposals from a test set or proposals generated during a real-world deployment).

Methods such as deformable part-based models and decision forests were the main approaches used for VRU detection until recent years \cite{benenson2014ten}. Artificial neural networks (ANNs) form the basis for many of the most successful detection algorithms which have been developed since. ANNs are loosely based on biological neural systems and are comprised of individual perceptrons which are interconnected. Deep learning (DL) \cite{goodfellow2016dl} is a ML paradigm which emerged relatively recently, encompassing a broad range of techniques based on ``deep'' ANNs (i.e. ANNs with multiple hidden layers of perceptrons). DL architectures allow complex concepts to be built using simpler concepts \cite{goodfellow2016dl}, e.g. a person is made out of body parts, body parts are made out of contours and edges, contours and edges are detected in arrays of raw pixel input.
Examples of DL-based detection algorithms which have achieved excellent performance at VRU detection from images in recent years include Fast-RCNN \cite{girshick2015fast}, Faster R-CNN \cite{ren2015faster}, R-FCN \cite{dai2016r}, YOLO \cite{redmon2016you} and SSD \cite{liu2016ssd}. Algorithms such as Pose-RCNN \cite{braun2016pose} have also been developed to leverage both image and lidar data concurrently.

\section{Open challenges and\\ promising future research directions}
\label{sec:openchallenges}

\subsection{Small \& partially occluded objects}
Small or partially occulded VRUs remain among the most challenging objects for AV perception systems to detect accurately. Stixel-based proposals are efficient, but need to be improved for these cases; future work should include careful tuning of stixel and proposal parameters (e.g. stixel width and segmentation costs) to address these challenges \cite{flohr2018vulnerable}.

\subsection{Detection of VRU gestures}
Future AVs will need to understand at least a minimal set of gestures made by humans, e.g. in cases where a cyclist is signalling a change of direction, or when a construction worker or police officer is manually directing traffic during construction works or at the scene of an accident. A first step in this direction would be to train AV perception systems to recognise individuals who could potentially give hand gestures, and then further classify these as VRUs giving advisory gestures (for cyclists) or directions which must be obeyed (for manual traffic control).

This presents difficulties due to differences in uniform styles across countries, as well as similarities in appearance when compared to other VRUs. Furthermore, high-visibility clothing is not exclusively worn by cyclists, police officers or construction workers, so extreme caution must be taken not to misinterpret hand movements of passers-by as genuine advisory or traffic control gestures. One possible solution in the medium term is for police and construction workers to have mobile transmitters which could communicate the need to hand back control to a human driver when manual gestures are in use, although this would not be an acceptable solution for a level 5 AV.

\subsection{False positives}
Further reducing the rate of false positives (i.e. detection of an object when one is in fact not present) remains an open challenge for computer vision researchers. VRU detection presents many opportunities for false detections to happen, e.g. due to reflections, clothes displayed in shop windows and large advertisements in the scene background which feature people. Some false positives may be eliminated by taking advantage of known scene geometry constraints (e.g. pedestrians or cyclists should be on the ground plane) \cite{braun2018EuroCity}. Object tracking between frames can also reduce the rate of false positives \cite{braun2018EuroCity} as reflections present in one frame may not be present in the next.

\subsection{Modelling and predicting VRU behaviour}
Humans drivers naturally predict the future movements of VRUs, and are able to anticipate events such as a pedestrian crossing the road suddenly. Developing models of VRU behaviour using ML techniques is an important direction for future work. Once a VRU has been detected and localised with the correct orientation, predictions of the future movements of VRUs could be integrated into the ego vehicle's path planning algorithm. Recent research \cite{goldhammer2018intentions} recorded the movement of VRUs at an intersection and learned models for predicting VRU trajectories based on their current trajectory. Future work could adopt such models for on-board VRU behaviour predicton in AV systems.

\subsection{Bayesian neural networks}
\begin{figure}[h]
\centering
\includegraphics[width=0.6\linewidth]{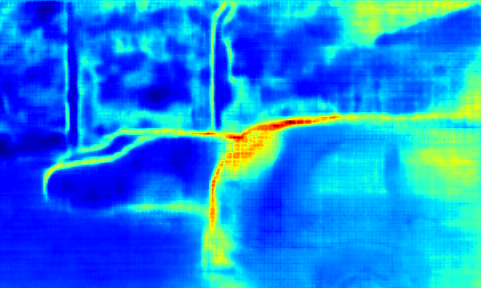}
\caption{Sample uncertinty map generated by a BDL approach. Note how higher uncertainty values are shown on difficult surfaces, such as vehicle windows. Figure reproduced from \cite{kendall2017uncertainties}.}
\label{fig:BDL}
\end{figure}
Current DL algorithms can effectively learn complex mappings between high-dimensional input data and a given set of outputs. However, a major shortcoming of these approaches is that they do not have any information about the uncertainty associated with a particular mapping. Bayesian deep learning (BDL) is a promising field which could address this shortcoming, while potentially achieving state-of-the-art-results \cite{kendall2017uncertainties}. BDL combines ideas from DL and Bayesian probability theory, so that an uncertainty value is associated with each object detection in each region of an image. A sample uncertainty map generated by a BDL algorithm is shown in Fig. \ref{fig:BDL}. Having an uncertainty value available may be useful to help to avoid false positives, an issue which was identified above as a major challenge for VRU detection systems. However, BDL is currently too computationally expensive to allow real-time inference on AV hardware (experiments in \cite{kendall2017uncertainties} ran slowly despite using an expensive and powerful NVIDIA Titan X GPU). Therefore, future work should investigate methods to speed up inference using this promising technique, so that it will become computationally viable for deployments on AVs. 

\subsection{Neuroevolution}
Typical DL algorithms make use of backpropagation to learn the network weights. One alternative is to use a population-based genetic algorithm (GA), where the weights of the network are encoded in genes. The GA then uses operators inspired by biological evolution such as selection, crossover and mutation to create new genes, and thus new sets of network weights. This alternative to backpropagation has gained popularity again recently, following successes such as Uber AI Labs' demonstration that GAs are a competitive alternative to backpropagation when training deep ANNs for reinforcement learning settings \cite{such2017deep}.

One advantage of GA-based methods is that they scale extremely well across large numbers of cores (tests were conducted across up to 720 CPUs in \cite{such2017deep}), as each solution in a new generation may be tested on its own core, allowing many candidate solutions to be evaluated at once. GAs are also effective at escaping local optima, especially quality-diversity algorithms \cite{pugh2016quality} such as novelty search. Neuroevolution could feasibly be used when training deep ANNs for object detection, potentially improving training speed and reducing the time needed to develop and test new algorithms for VRU detection.

\subsection{V2X communication}
As with human drivers, even the best AV perception systems which are based on line-of-sight will not be able to detect VRUs that are fully occluded (e.g. behind another vehicle, a building or another VRU). Future AVs are likely to be connected to other vehicles (V2V), to infrastructure (V2I) and possibly even to VRUs (V2P). Prototype V2V systems such as Valeo XtraVue\footnote{\url{https://youtu.be/F4-wWfCcyK4}} allow sensor data to be transmitted between adjacent vehicles, which could reduce the number of objects in a scene which are occluded by other vehicles. The fusion of data from on-board sensors with data received from V2P communications for improved pedestrian detection is also currently being investigated (see e.g. \cite{merdrignac2017fusion}). The increased availability of information resulting from these methods is likely to significantly improve VRU detection capabilities. 

\section{Conclusion}
\label{sec:conclusion}
This paper presented an overview of recent research and the current state-of-the-art in VRU detection, followed by a discussion of the main open challenges and promising future research directions. AV systems are still very much at the prototype stage and are unlikely to secure regulatory approval for widespread deployment unless they are at least as safe as human drivers. The most fundamental requirement for safe AVs is a fast and accurate environment perception system. Computer vision engineers must pay special attention to VRUs as they are among the most difficult objects to detect, as well as having a high risk of death or serious injury in the event of a collision. Focusing research efforts on promising directions such as those listed in Section \ref{sec:openchallenges} is essential if the performance of AV perception systems for VRU detection is to meet or exceed that of human drivers in the coming years.

\bibliographystyle{IEEEtran}
\bibliography{bibliography}

\end{document}